\documentclass{article}

\usepackage{arxiv}

\usepackage[utf8]{inputenc} 
\usepackage[T1]{fontenc}    
\usepackage{url}            
\usepackage{booktabs}       
\usepackage{amsfonts}       
\usepackage{nicefrac}       
\usepackage{microtype}      
\usepackage{lipsum}
\usepackage{graphicx}
\usepackage{orcidlink}
\usepackage{comment}
\usepackage{graphics} 

\usepackage{amsmath} 
\usepackage{algorithm}
\usepackage{algorithmic}
\usepackage{wrapfig}
\usepackage{sidecap}
\usepackage{caption}

\usepackage{subcaption}

\definecolor{VUB_blauw}{rgb}{0.1529, 0.2667, 0.5529}
\usepackage{hyperref}       
\hypersetup{
    colorlinks,%
    citecolor=VUB_blauw,%
    filecolor=VUB_blauw,%
    linkcolor=VUB_blauw,%
    urlcolor=VUB_blauw
}
\graphicspath{ {./images/} }
\newcommand{\customCor}[1]{%
  \includegraphics[height=1em]{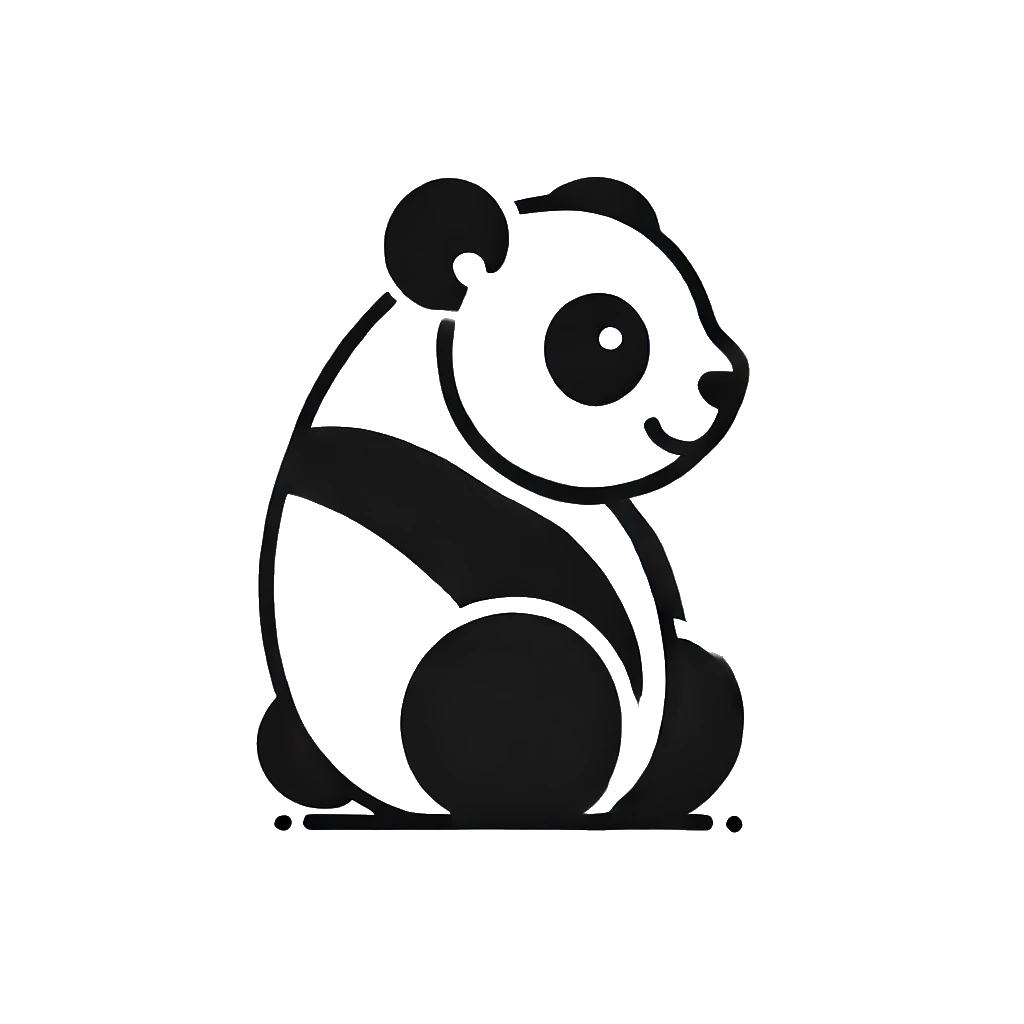} #1%
}
\newcommand{\Lim}[1]{\raisebox{0.5ex}{\scalebox{0.8}{$\displaystyle \lim_{#1}\;$}}}

\AddToHook{shipout/background}{%
  \ifnum\value{page}=1 
    \pagenumbering{Roman} 
    \setcounter{page}{1} 
  \fi
  \ifnum\value{page}=2 
  \pagenumbering{arabic} 
  \setcounter{page}{2} 
  \fi
}

\title{Model-Agnostic Solutions for Deep Reinforcement Learning in Non-Ergodic Contexts}
\runningtitle{Deep RL solutions for Non-Ergodic Dynamics}

\author{
  Bert Verbruggen\textsuperscript{1,\customCor{ }} \\ 
  \orcidlinkc{0000-0001-9776-2420} \\
  \And
  Arne Vanhoyweghen\textsuperscript{1,2} \\ 
  \orcidlinkc{0000-0003-0103-4715} \\
  \And
  Vincent Ginis\textsuperscript{1,3} \\ 
  \orcidlinkc{0009-0001-5086-2704} \\
  \and
  \textsuperscript{1}Data Analytics Lab, Vrije Universiteit Brussel, 1050 Brussel, Belgium \\ 
  \textsuperscript{2}Mobilise, Vrije Universiteit Brussel, 1050 Brussel, Belgium \\ 
  \textsuperscript{3}School of Engineering and Applied Sciences, Harvard University, Cambridge, Massachusetts 02138, USA
}
    
\begin{document}

\maketitle
\renewcommand{\thefootnote}{} 
\footnotetext{\includegraphics[height=1em]{panda2.png} Corresponding author: bert.verbruggen@vub.be}
\renewcommand{\thefootnote}{\arabic{footnote}} 
\thispagestyle{plain} 

\begin{abstract}
    Reinforcement Learning (RL) remains a central optimisation framework in machine learning. Although RL agents can converge to optimal solutions, the definition of ``optimality'' depends on the environment's statistical properties. The Bellman equation, central to most RL algorithms, is formulated in terms of expected values of future rewards. However, when ergodicity is broken, long-term outcomes depend on the specific trajectory rather than on the ensemble average. In such settings, the ensemble average diverges from the time-average growth experienced by individual agents, with expected-value formulations yielding systematically suboptimal policies. Prior studies demonstrated that traditional RL architectures fail to recover the true optimum in non-ergodic environments. We extend this analysis to deep RL implementations and show that these, too, produce suboptimal policies under non-ergodic dynamics. Introducing explicit time dependence into the learning process can correct this limitation. By allowing the network’s function approximation to incorporate temporal information, the agent can estimate value functions consistent with the process’s intrinsic growth rate. 

    This improvement does not require altering the environmental feedback, such as reward transformations or modified objective functions, but arises naturally from the agent’s exposure to temporal trajectories. Our results contribute to the growing body of research on reinforcement learning methods for non-ergodic systems.
\end{abstract}


\section{Introduction}
Reinforcement Learning (RL) is a powerful training method for agents to learn policies that optimise actions in decision-making contexts~\cite{sutton2018reinforcement}. An agent is set in a virtual environment with states $s_{t} \in \mathcal{S}$, and available actions $a_{t} \in \mathcal{A}$ for a time-step $t \in \mathbb{N}$. In traditional contexts, the state-action space, the set of all possible state-action pairs, is represented in a table, and a value function $V(s_{t})$ is learned during training. This function determines a reference value for choosing a specific action from a given state. To find this function, an optimisation is implemented based on the Bellman Equation,
\begin{equation}
    V(s) = \mathbb{E}\left[ r_{t+1} + \gamma V(s_{t+1}) \mid s_{t} = s\right].
    \label{eq:bellman}
\end{equation}
This equation has an explicit dependence on the expected value ($\mathbb{E}[R]$) of the rewards $R$ and resides at the core of traditional update rules such as SARSA~\cite{rummery1994line}, Q-learning~\cite{watkins1992q} or the Monte Carlo (MC) methods~\cite{sutton2018reinforcement}.

Despite the effective implementations and results of RL over the past years, its reliance on expected values implicitly assumes ergodicity and restricts the standard implementation of RL to ergodic contexts. In an ergodic dynamic, the expected value is equal to the time growth of an agent, as indicated by Equation~\ref{eq:ergodicity}. When this equality does not hold, we say ergodicity is broken. For a generic optimiser function $f(\omega)$, with $f:\Omega \to \mathbb{R}$ an integrable observable evaluated along a trajectory $\omega$, this equation can be written as
\begin{equation}
    \Lim{T \to \infty} \frac{1}{T} \int^{T}_{0} f\left(\omega\left(t\right)\right) dt = \int_{\Omega} f\left(\omega\right) P\left(\omega\right) d\omega.
    \label{eq:ergodicity}
\end{equation}
In non-ergodic contexts, however, the optimal policy cannot be accurately inferred from expected values alone. Recent work~\cite{verbruggen2025reinforcement} demonstrated this effect in multiplicative dynamics, where path dependence requires optimisation with respect to time-average growth rather than ensemble averages. Prior studies focused primarily on tabular RL with finite, discrete action spaces. In the present work, we extend this analysis to Deep Reinforcement Learning (DRL) by employing function approximators to generalise over continuous and higher-dimensional spaces. Although neural networks, as function approximators, are designed to capture complex dependencies in the data, our findings show that this capability does not resolve the underlying limitation: when the learning signal is defined in terms of expected values, DRL agents remain unable to identify the true time-optimal policy.

\section{Ergodicity in Reinforcement Learning Research}
The concept of ergodicity has broad relevance across scientific disciplines, from physics to social sciences. In the present work, we examine its implications within agent-based modelling, particularly in Economics and Finance, where a seminal contribution was made by~\cite{peters2019ergodicity}. In this work, the author formalises the notion of ergodicity breaking and demonstrates its importance for understanding economic dynamics. The author's initial critique focuses on the widespread use of ensemble averages in behavioural economics and the lack of motivation for the choice of specific utility functions. Peters illustrates this by investigating the multiplicative dynamic, a compounding process at the heart of many economic applications, in which wealth grows and shrinks with relative increments. Moreover, Peters shows that the multiplicative dynamic is non-ergodic and therefore well described by the time-average growth rate, but not by its expected value~\cite{peters2016evaluating}. Finally, the author demonstrates that, with the right transformation, non-ergodic dynamics can be rendered ergodic, allowing the expected values of the transformed variables to meaningfully represent long-term outcomes. This transformation, often referred to as the ergodic transform~\cite{hulme2023reply}, bears similarities to the notion of a utility function. However, whereas utility functions arise from idiosyncratic preferences, the ergodic transform follows from the objective structure of the underlying dynamics. 

Subsequent studies in Ergodicity Economics have demonstrated the far-reaching implications of the ergodic assumption across the behavioural and computational sciences. This ranges from findings that human decision-making, often labelled as irrational under classical expected-utility theory, can be re-interpreted as rational when evaluated through time-average growth~\cite{meder2021ergodicity, vanhoyweghen2022influence, vanhoyweghen2023human, VANHOYWEGHEN2025101663}, to analyses showing how non-ergodic dynamics can promote cooperation and collective stability in social and economic systems~\cite{peters2023insurance, peters2022ergodicity, vanhoyweghen2025redistributive,fant2023stable}. The influence of ergodicity-breaking extends further to reinforcement learning, where agents trained on expected-value criteria have been shown to develop sub-optimal policies in non-ergodic environments~\cite{verbruggen2025reinforcement, baumann2023reinforcement, sheng2025beyond}.

When combining decision-making and risk sensitivity, we often encounter RL and Markov decision processes (MDPs). Foundational work introduced risk into Markov Decision Processes (MDPs) as a selective parameter for value iteration and policy iteration~\cite {howard1972risk}. By embedding risk-sensitivity as a parameter in the objective function of the exponential expected utility, a similar notion to the ergodic transform was introduced. An alternative introduction to the value function is considered by~\cite{heger1994consideration}, which incorporates the variance of the return into the optimisation. This formulation is often referred to as the first risk-aware RL formulation. By including variance in the objective function, it diverges from optimising the expected value alone and also focuses on variance minimisation. In a random multiplicative process, whose central tendency depends on both the expected return and its variance, this approach can be interpreted as approximately optimising the time-average growth rate. 
Similar conceptual introductions of parameters have been studied, using variance, entropy or other definitions to modulate the effects of the stochastic return signal as well~\cite{mihatsch2002risk, di2012policy, haarnoja2018soft}. 

A complexity analysis for mean-variance objectives, including variance-driven operators in the objective functions for agent-based modelling, shows that such objectives are non-additive and, as such, non-ergodic by design. Related to the specific use of actor-critic (AC) models,~\cite{prashanth2016variance} presents three approaches for incorporating variance into the optimisation of AC models and demonstrates their practical application in a dedicated experimental setting. 

A different approach from explicitely accounting for the variance in the value function, introduces the ergodic transform in the objective function during optimisation~\cite{bielecki1999risk}. For instance, using logarithmic utility directly resolves the broken ergodicity of the return signal in portfolio optimisation.
In recent research on RL under non-ergodic contexts,~\cite{baumann2023reinforcement} discusses how this transformation can be learned to make reward increments ergodic before computing the return signal. In recent work~\cite{sheng2025beyond}, the optimisation itself is altered to focus on growth rates rather than expected values. By replacing the arithmetic mean with the geometric mean as an objective, the agents learn to optimise average time growth through proper dynamic optimisation. 

A solution that effectively handles a non-ergodic context is to use an actual distribution in the objective function formulation, rather than relying on a single or several statistical parameters. This results in a statistical interpretation of the problem rather than a transformation of the objective itself~\cite{bellemare2017distributional}.

In a recent survey of risk-sensitive RL~\cite{biswas2023ergodic}, a concise categorisation of methods for Ergodic Risk-Sensitive Control (ESRC) also underscores the important contribution of our work, as it stands out from interventions in rewards or value functions.

All of the previous approaches rely on ex ante knowledge of the underlying dynamic and its corresponding ergodic transform. In contrast, we propose an alternative framework that does not alter the objective function algebraically but instead exploits two intrinsic properties: the path-dependent nature of non-ergodic dynamics and the finite-horizon character of reinforcement learning. Rather than training a large number of agents on single-step realisations of a stochastic process, our approach emphasises learning over repeated time trajectories, where compounding effects become observable. This design aligns the agent’s learning dynamics with the process’s intrinsic time-average growth, effectively embedding ergodic reasoning into the training structure itself. 
Even in the absence of a generalised Central Limit Theorem for random multiplicative processes~\cite{redner1990random}, the typical outcomes, i.e. those occupying the densest regions of the wealth distribution over time, dominate the experiences available to an RL agent given sufficient repetitions of the learning setup. This method has proven effective in prior work on traditional update rules~\cite{verbruggen2025reinforcement}, but lacks the validation in a contemporary context where RL training is improved by using neural networks as function approximators. Our work shows that increasing model complexity alone is insufficient to achieve proper optimisation when training agents in non-ergodic contexts, and that additional incentives, such as our repetition-based training for temporal dynamics, are required to formulate the optimal policy.

\section{Experiment design}
To investigate the link between architectural complexity and RL's ability to account for ergodicity-breaking, we reproduce the original two experiments proposed in~\cite{verbruggen2025reinforcement} in an updated environment that includes relevant DRL models. The current scope identifies the behaviour of traditional DRL agent implementations for decision-making in a context with multiplicative dynamics. To investigate the central role of the expected value in DRL contexts with path-dependent dynamics, we will adapt a similar experimental approach. By evaluating the policy in a traditional training context, we can identify the agent's default optimisation strategy. We will further extend the problems to an explicit time dimension to evaluate the proposed solution from prior work~\cite{verbruggen2025reinforcement}, now in the context of value function approximations. This means we enforce the agent's path dependence by repeating the agent's essential step multiple times. In doing so, the underlying temporal dynamic is exposed and can be learned by the agent during training. Our study focuses on the exemplary studies in a traditional RL context, a toy model for multiplicative dynamics and a portfolio assignment problem.

\subsection{Toy model for an MDP}
\label{sec:toymodel}
\begin{figure}
  \centering
  \begin{subfigure}{0.48\textwidth}
    \includegraphics[width=\linewidth]{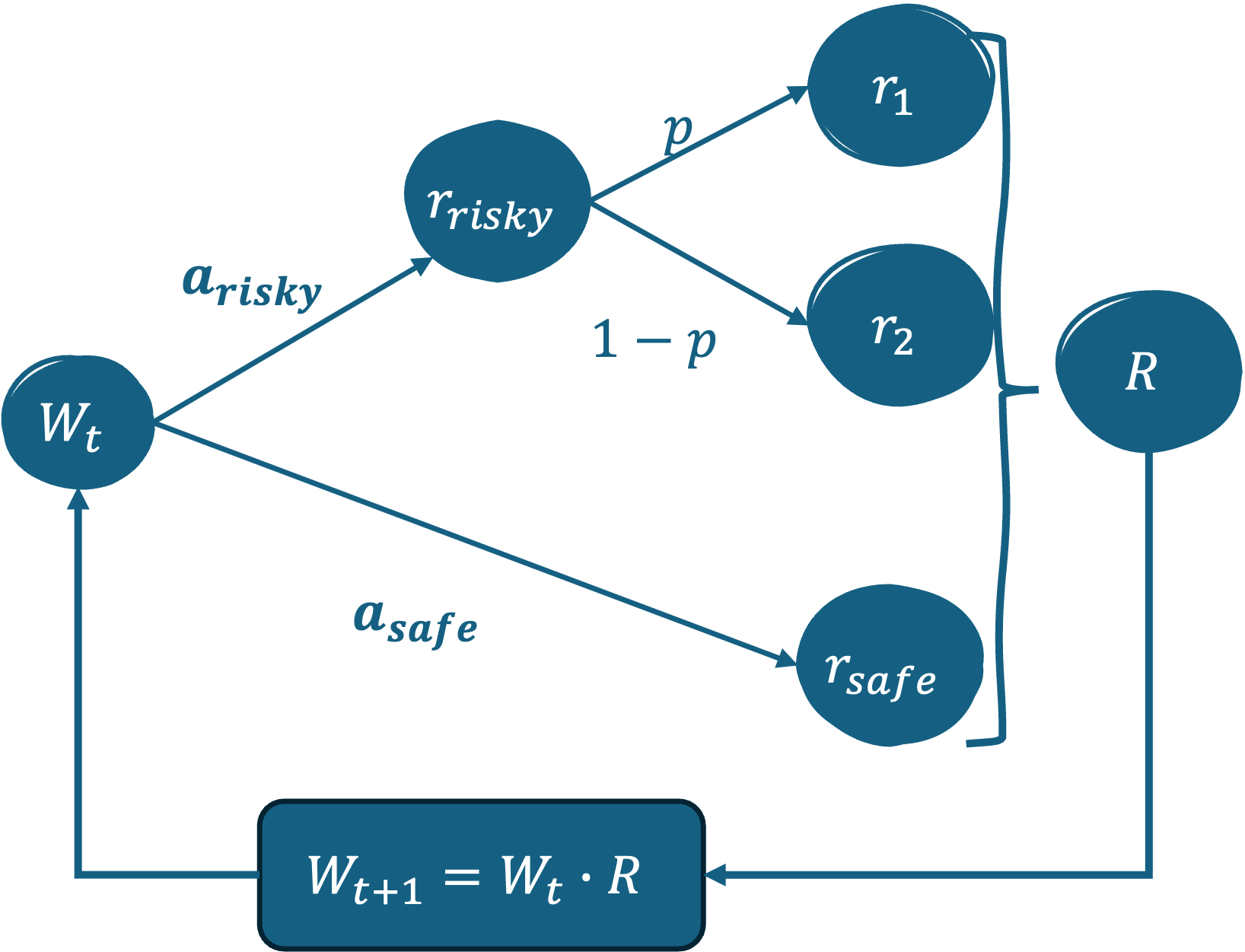}
    \caption{}
    \label{fig:toy_model}
  \end{subfigure}
  \hfill
  \begin{subfigure}{0.48\textwidth}
    \centering
    \includegraphics[width=\linewidth]{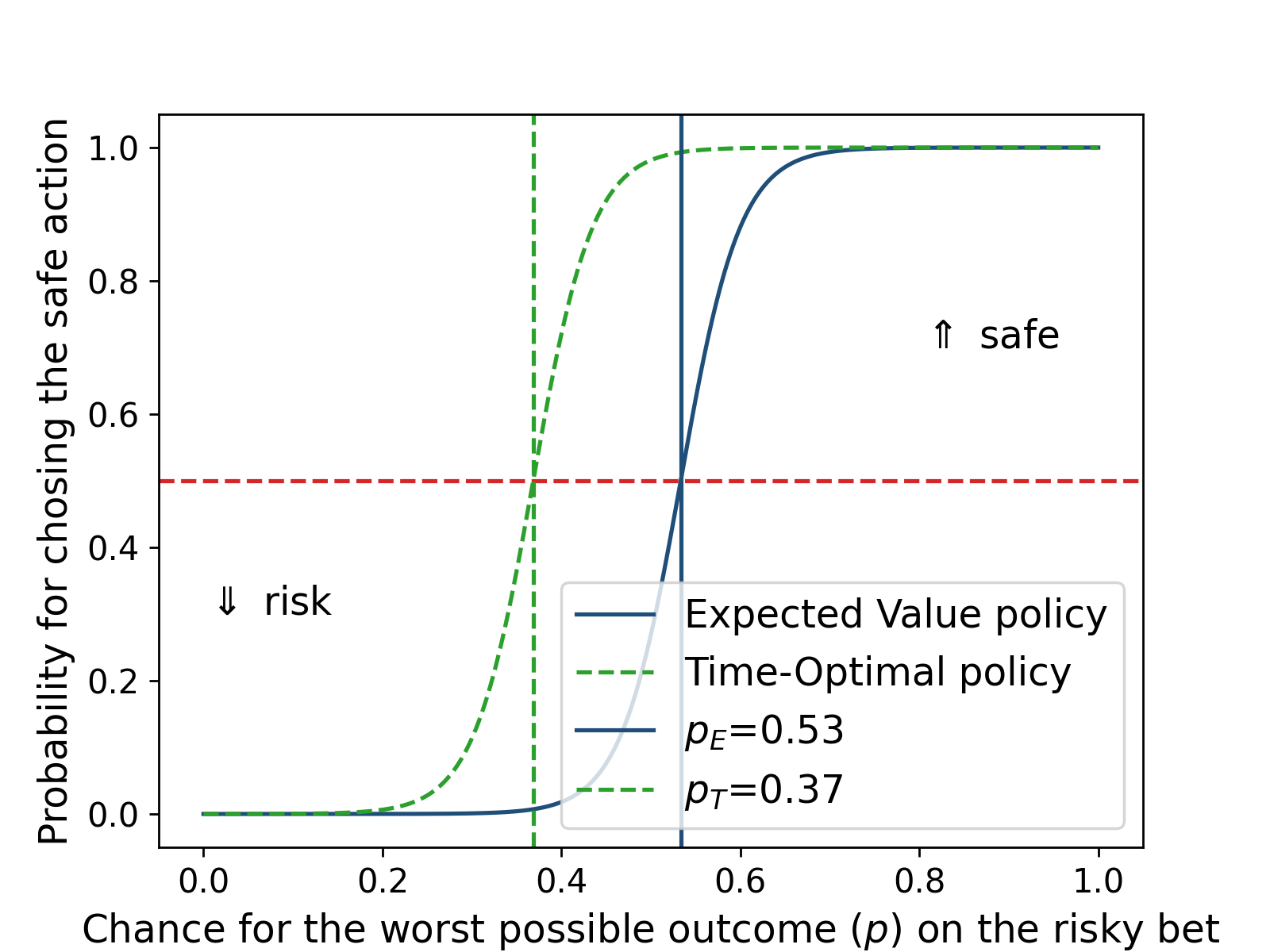}
    \caption{}
    \label{fig:genericpolicy}
  \end{subfigure}

  \caption{
  Illustration of our toy model for training an agent in a multiplicative dynamic~\subref{fig:toy_model}. An agent needs to decide which of two actions ($a_\text{safe}, a_{risky}$) is most beneficial for optimising its wealth, given a probability $p$ to receive the worst outcome on the risky action. Our proposed solution to manage the non-ergodic context uses repetitions in the action-selection, indicated by the return of the updated wealth to the state for the next time step. Illustration of the two different policies~\subref{fig:genericpolicy}. The blue solid function describes an optimisation based on expected values and predicts the indifference point $p_{E}$. The green dashed function simulates an agent optimising for growth rates, with the predicted indifference point at $P_{T}$.}
  \label{fig:toymodel_overal}
\end{figure}

To illustrate the learning dynamics in a controlled non-ergodic setting, we introduce a simple toy model of multiplicative wealth growth, shown in Figure~\ref{fig:toy_model}. In this setup, an agent is presented with two possible actions, $\mathcal{A} = (a_\text{safe}, a_\text{risky})$, and a single state $s_t$ for each time step. The agent must learn to optimise its expected wealth by choosing between these actions, depending on the probability $p$ of receiving the worst possible outcome from the risky option. An agent starts at time step $t$ with initial wealth $s_{t} = W_{t}$ and chooses between an action ($a_\text{safe}$), that results in a small but consistent multiplicative increase in its wealth with factor $r_\text{safe}$, and alternative action ($a_{risky}$) results in a stochastic reward $r_{risky}$ following a binomial distribution. The risky action can increase the agent's wealth with a larger factor $r_{2} > r_{1}$, but can decrease its initial wealth $r_{1} < 1$ with probability $p$. The resulting reward $R \in (r_{1}, r_{2}, r_\text{safe})$ is multiplied by the agent's initial wealth $W_{t}$ to provide the initial wealth for the next time step, 
\begin{equation}
    W_{t+1} = R \cdot W_{t}.
    \label{eq:wealthupdate}
\end{equation}
Writing this with the explicit outcomes of the actions, subject to the binomial process under the risk action, we find that the agent has three possible outcomes.
\begin{align}
    \label{eqn:coin_toss_temp_train}
    W_{t+1} = \begin{cases}
    r_{\text{safe}} \cdot W_{t} &\text{ if } a_{t} = a_\text{safe} \text{,}\\
    r_{1} \cdot W_{t} &\text{ if } a_{t} = a_\text{risk} \text{, w.p. } p, \\
    r_{2} \cdot W_{t} &\text{ if } a_{t} = a_\text{risk} \text{, w.p. } (1-p), 
    \end{cases}
\end{align}
In traditional RL experiments, multiple agents are trained independently, each learning the optimal decision for a given probability $p$ of receiving the worst outcome under the risky action.

Traditional RL pipelines require a table of Q-values for each state-action pair to estimate their value within the optimisation problem. The ``optimiser'' now selects the action that returns the highest Q-value for the current state, $\max_{a}{Q(s,a)}$, from the table. In our first experimental setup, we replace this tabular representation, used in ~\cite{verbruggen2025reinforcement} with a neural-network-based function approximator.  This transition from traditional RL to deep reinforcement learning (DRL) enables the agent to handle larger state spaces and more complex environments. These function approximators replace the value selection problem with a value function which the model needs to optimise. We use an implementation of Deep Q-learning~\cite{mnih2015human}, which utilises the Bellman equation to compute the loss backpropagated to the neural network during training. We illustrate the experimental design for the agents using a DQN in Algorithm~\ref{alg:DQN}. Moreover, we transform the Multi-Armed Bandit (MAB) setup from~\cite{verbruggen2025reinforcement} into a Markov Decision Process (MDP) by including the current wealth, $W_{t}$, in the state at time step $t$.

\begin{algorithm}[t]
\caption{DQN implementation for the toy model}
\label{alg:DQN}
\begin{algorithmic}[1]
\STATE Given: probabilities $\mathcal{P}$; agents $N$; episodes $E$; rounds per episode $M$; action set $\mathcal{A}$ 
\STATE Q-approximator: feed-forward net $Q_{\theta}(s,\cdot)$ (input: wealth $w$; output: $|\mathcal{A}|$ Q-values); loss: SmoothL1; optimizer: Adam$(\eta)$
\FOR{each $p \in \mathcal{P}$}
  \FOR{$j = 1$ \textbf{to} $N$}
    \FOR{$e = 1$ \textbf{to} $E$}
      \STATE Reset environment: $W \gets 1$; $t \gets 0$
      \WHILE{$t < M$}
        \STATE Form state $s \gets W$; compute $Q \gets Q_{\theta}(s,\cdot)$
        \STATE Select action $a$ by $\varepsilon$-greedy over $Q$
        \STATE Sample from the  environment with $(p,a)$ to obtain reward $r_t$ and next wealth $W'$
        \STATE Set $s' \gets W'$; append transition $(s,a,r_t,s')$ to buffer $\mathcal{D}$
        \STATE $W \gets W'$; $t \gets t+1$ 
      \ENDWHILE
        \STATE \textbf{Evaluation at fixed $p$:} set $\varepsilon \gets 0$; form policy from $Q_{\theta}$
        \STATE record $\pi_{\text{safe}}$ for buffer $\mathcal{D}$ as target
        \IF{$|\mathcal{D}| \ge B$}
          \STATE Build mini-batch of size $B$ from recent transitions in $\mathcal{D}$
          \STATE For each $(s_i,a_i,r_i,s'_i)$ in batch:
          \STATE \hspace{1em} $y_i \gets r_i + \gamma \max_{a'} Q_{\theta}(s'_i,a')$
          \STATE \hspace{1em} $\hat y_i \gets Q_{\theta}(s_i,a_i)$
          \STATE Accumulate loss $L \gets \frac{1}{B}\sum_{i=1}^{B}\text{SmoothL1}(\hat y_i, y_i)$
          \STATE Backpropagate $L$ and update $\theta$ with Adam
        \ENDIF
    \ENDFOR
  \ENDFOR
  \STATE Aggregate over runs to obtain $\widehat{\pi}_{\text{safe}}(p)$
\ENDFOR
\STATE Return $\{\widehat{\pi}_{\text{safe}}(p): p \in \mathcal{P}\}$
\end{algorithmic}
\end{algorithm}
The DQN uses the agent's current wealth at time step $t$, $W_{t}$, as input to the network model. After the forward pass, the model returns a preference for each action, represented by two output nodes. By employing a decaying epsilon-greedy approach for action selection, we enable the agent to explore effectively. The environment then returns the received reward and updated wealth for the next step. We collect these transitions for a full episode and gather several episodes in a buffer before updating the network weights based on the loss calculated from the current model preferences and the past trajectories in the buffer. The resulting policy is derived from the different preferences the trained network assigns to individual fixed probabilities of receiving the worst outcome in the bet. The experimental setup from a training perspective is presented in Appendix~\ref{appendix_dqn}, with model and training parameters.

Next, we can compare the policy of agents trained on the toy model for a fixed set of environment definitions, the number of repetitions ($M$), and the probability of receiving the worst outcome from the risky action ($p$). We train a fixed number of independent agents ($N$) to learn the optimal action in this environment. By repeating this process for different values of the parameter $p$, we can evaluate the policy of the agents trained in our experiment and compare it to a prediction based on each of the optimisers. Under an expected-value-based optimisation, the agent's behavioural shift from risk seeking to risk aversion occurs at a predictable value of the parameter $p$. The indifference point can be calculated using the expected reward for each action,
\begin{align}
    \mathbb{E}_\text{risk}\left[W_{t+1}\right] &= \left(p \cdot r_{1} + \left( 1-p \right) \cdot r_{2}\right) \cdot W_{t}
    \label{Expected_risk} \\
    \mathbb{E}_\text{safe}\left[W_{t+1}\right] &= r_\text{safe} \cdot W_{t}.
    \label{Expected_Safe}
\end{align}
This is called the indifference point $p_{E}$ and can be determined by setting both expected rewards equal to each other. 
\begin{align*}
    \mathbb{E}_\text{safe}\left[W_{t}\right] &= \mathbb{E}_\text{risk}\left[W_{t} \right] \\
    \Rightarrow r_\text{safe} &= p_{E} \cdot r_{1} + \left(1-p_{E} \right) \cdot \left(r_{2} \right) \\ 
    \Rightarrow p_{E} &= \frac{r_\text{safe}-r_{2}}{r_{1} - r_{2}} \\
\end{align*}
A similar derivation can also be used to predict the indifference point based on the process's growth rate. For this, we use the formulation of the binomial process to find the expected rewards.
\begin{align*}
    r_\text{safe} &= (r_{1})^{p_{T}} \cdot (r_{2})^{(1-p_{T})} \\ 
    \Rightarrow \ln{(r_{\text{safe}})} &= \ln{\left[(r_{1})^{p_{T}} \cdot (r_{2})^{(1-p_{T})}\right]} \\ 
    \Rightarrow \ln{(r_\text{safe})} &= p_{T} \cdot \ln{(r_{1})} + (1-p_{T}) \cdot \ln{(r_2)} \\
    \Rightarrow p_{T} &= \frac{\ln{\left(r_\text{safe}\right)}-\ln{\left(r_{2}\right)}}{\ln{\left(r_{1}\right)} - \ln{\left(r_{2}\right)}}
\end{align*}
An agent will start off always preferring the risky action when the probability of receiving the worst outcome is zero, $p=0$. When the probability of receiving the worst outcome increases, so too does the agent's likelihood of selecting the safe action. This is up to the indifference point where each action is equally likely to be selected. From then on, the agent will predominantly prefer the safe action until it only considers this action. This dynamic can be simulated using a sigmoid function, with the parameter $p$ as its input,
\begin{equation}
    \sigma(p) = \frac{1}{1+e^{-k(p-p_{0})}}.
    \label{Equation_sigmoid}
\end{equation}
Here $p_{0}$ is the indifference point at which the agent switches from predominantly taking the risky action to the safe action, the abruptness of which is given by the scaling parameter $k$.\\

In this experiment, we will evaluate a regression of the agent's probability for taking the safe action as a function of this parameter $p$ and compare the location of the indifference point. As illustrated in Figure~\ref{fig:genericpolicy}, the location of this indifference point indicates whether an agent optimises its decision based on the expected value prediction $p_{E}$ or on a time-growth perspective $p_{T}$.
\subsection{Portfolio Assignment Optimisation}
\label{sec:pamodel}
Portfolio assignment is a popular problem in financial contexts, where you need to decide how to allocate your wealth across assets with varying multiplicative returns. Because of the multiplicative, i.e. compounding, nature of the returns, this problem is inherently non-ergodic. Consequently, its optimal solution is not described in terms of expected return but rather by the Kelly objective~\cite{kelly1956new},
\begin{equation*}
    g(f) = \mathbb{E}\left[\log{((1-f^{*}) + f^{*} \cdot R) }\right],
    \label{eq:kelly_objective}
\end{equation*}
where $R$ represents the multiplicative return from a portfolio and $f^{*}$ denotes the optimal fraction of your wealth to invest in the portfolio to maximise the total return. A simplified setup of this problem can be illustrated by a construction similar to the toy model. An agent now needs to propose a fraction ($f^{*}$) of its wealth to invest in a portfolio, described by a stochastic bet with an underlying binomial process. This portfolio has two possible outcomes: an increase in invested wealth with probability $p$ and a loss via a multiplicative decrease otherwise. The fraction of non-invested wealth remains constant. 
\begin{figure}
  \centering
  \begin{subfigure}{0.48\textwidth}
    \includegraphics[width=\linewidth]{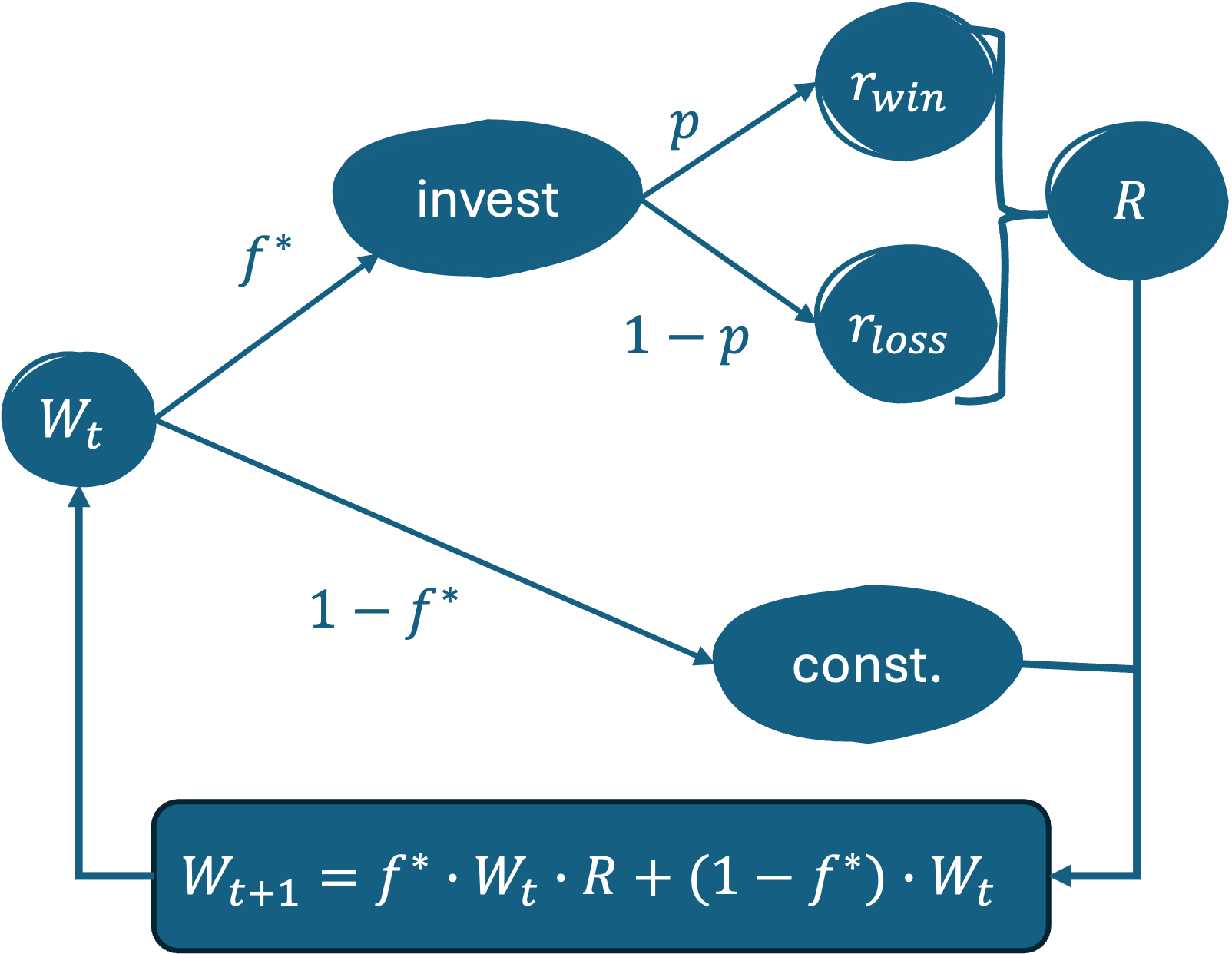}
    \caption{}
    \label{fig:kelly_setup}
  \end{subfigure}
  \hfill
  \begin{subfigure}{0.48\textwidth}
    \centering
    \includegraphics[width=\linewidth]{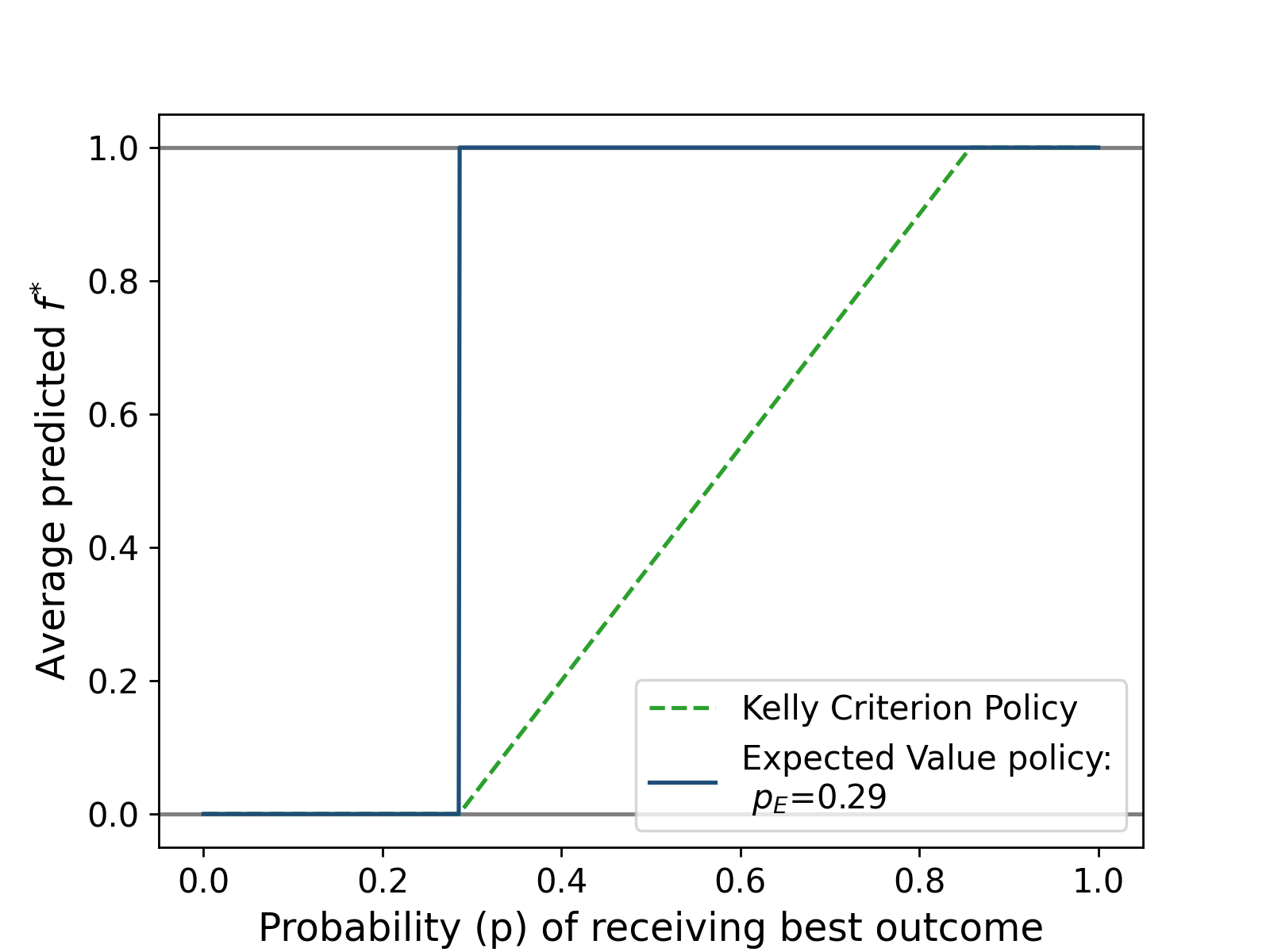}
    \caption{}
    \label{fig:kellygeneric}
  \end{subfigure}

  \caption{
  Illustration of the portfolio assignment problem, where the agent needs to find the optimal fraction ($f^{*}$) of its wealth to invest in a portfolio~\subref{fig:kelly_setup}. Illustration of an expected value policy, solid blue curve, and the optimal policy following the Kelly criterion, dashed green curve, for a policy assignment optimisation problem shown in~\subref{fig:kellygeneric}.}
  \label{fig:kellyIllustration}
\end{figure}
The optimal fraction can be derived and depends on the probability of receiving the best outcome in the portfolio.
In our experimental setup, the Kelly objective from Equation~\ref{eq:kelly_objective} can be rewritten using the exact binomial outcome as,
\begin{equation}
    g(f) = p \log{\left(1 + f(r_{\text{win}} -1)\right)} + (1-p) \log{\left(1+ f(r_{\text{loss}}-1) \right)}
    \label{eq:kelly_explicit}
\end{equation}
By setting the derivative equal to zero, $g'(f) = 0$ we can find the optimum of the objective function. This simplifies to solving,
\begin{align}
    \label{eq:pa_kelly}
    &\frac{d g(f)}{df} = 0 \\
    &\Leftrightarrow \frac{p(r_{\text{win}} - 1)}{1 + f (r_{\text{win}} - 1)} + \frac{(1-p) (r_{\text{loss}} -1)}{1 + f(r_{\text{loss}} - 1)} = 0 \\
    &\Leftrightarrow f^{*} = \frac{p(r_{\text{win}} - r_{\text{loss}}) + r_{\text{loss}} - 1}{(r_{\text{win}} - 1)(1 - r_{\text{loss}})}
\end{align}

In contrast, an agent optimising for expected value will adopt an all-or-nothing strategy, investing all their money in the portfolio once its expected return is greater than one, i.e. 
\begin{equation*}
\mathbb{E}[W_{t+1}] = \mathbb{E}[R] \cdot W_{t} \geq W_{t}.
\end{equation*}
We can compute the theoretical switching point $(p_{E})$ for a portfolio with rewards ($r_{win},r_{loss}$) and a non-invested wealth that remains constant, by comparing the expected wealth for each decision. When these expectation values are equal, we can extract the probability of winning as the prediction of where the agent will invest its wealth in the portfolio. This time, there is no optimal fraction to determine, as the strategy is either to invest all your wealth in the portfolio or keep everything constant, depending on the probability of winning from the investment. As shown by~\cite{kelly1956new}, this policy will be suboptimal.
\begin{align}
    \label{eq:pa_exp}
    \mathbb{E}[W_{t+1}] &= f \cdot W_{t} \cdot p(r_{\text{win}} + (1-p)r_{\text{loss}}) + (1 - f) \cdot W_{t} \\
    \Rightarrow \frac{\partial \mathbb{E}[W_{t+1}]}{\partial f} &= 0 \\
    \Leftrightarrow 0 &= W_{t} \cdot (p(r_{\text{win}} - r_{\text{loss}}) + r_{\text{loss}}) - W_{t} \\
    \Leftrightarrow p_{E} &= \frac{1- r_{\text{loss}}}{r_{\text{win}} - r_{\text{loss}}}
\end{align}

Resulting in the following fractions invested by the expected value optimising agent.
\[
f^{*} = 
\begin{cases}
1, & \text{w.p. } p\geq p_{E}, \\
0, & \text{w.p. } p\leq p_{E}.
\end{cases}
\]

We now introduce a generalisation of the portfolio assignment problem from~\cite{verbruggen2025reinforcement} into a DRL context. This problem cannot be modelled using the DQN implementation discussed before, as we want the agent to predict the fraction of the wealth to invest in the portfolio from a continuous action space, $f^{*} \in [0,1]$. A continuous action space requires a different approach, modelling the agent as an actor-critic~\cite {sutton2018reinforcement,lillicrap2015continuous}. These models use two separate neural networks: an actor, which learns the policy and selects actions, and a critic, which estimates the value function and provides a reference for the current state-action's value within the environment, thereby calculating the error in the actor's actions. This allows for continuous action spaces and is a logical artificial model for training agents on the portfolio assignment problem. Using a standard off-the-shelf actor-critic, we can implement the portfolio assignment problem as illustrated in~\ref{fig:kelly_setup} to estimate the Kelly criterion in a DRL context, as delineated in Algorithm~\ref{alg:AC_simple}. 
\begin{algorithm}
\caption{Actor--Critic for portfolio assignment (continuous action space)}
\label{alg:AC_simple}
\begin{algorithmic}[1]
\STATE Given: probabilities $\mathcal{P}$; agents $N$; episodes $E$; rounds per episode $M$
\STATE Networks: actor $\pi_{\theta}(f \mid s)$ outputs a distribution over fractions $f \in [0,1]$; critic $V_{\psi}(s)$ outputs a scalar value estimate
\FOR{each probability $p \in \mathcal{P}$}
  \FOR{$i = 1$ \TO $N$}
    \FOR{$e = 1$ \TO $E$}
      \STATE set state $s \gets W_0$ 
      \STATE Actor proposes fraction $f$
      \FOR{$m = 1$ \TO $M$}
        \STATE Apply $(p,f)$ in environment $\Rightarrow$ updated wealth $W_{m+1}$
      \ENDFOR
      \STATE Compute episode reward $R \gets W_M - W_0$ 
      \STATE Compute return $G \gets R$ 
      \STATE Compute advantage $A \gets G - V_{\psi}(s)$
      \STATE policy loss $L_{\pi} \gets -\,\log \pi_{\theta}(f \mid s)\,A$
      \STATE value loss $L_{V} \gets (G - V_{\psi}(s))^2$
      \STATE Update actor: $\theta \leftarrow \theta - \eta_{\pi}\,\nabla_{\theta} L_{\pi}$
      \STATE Update critic: $\psi \leftarrow \psi - \eta_{V}\,\nabla_{\psi} L_{V}$
    \ENDFOR
    \STATE Evaluate policy at fixed $p$ using deterministic actor output
  \ENDFOR
  \STATE Aggregate evaluation results across agents
\ENDFOR
\STATE Return summary of learned policies $\{\widehat{f}^{\ast}(p) : p \in \mathcal{P}\}$
\end{algorithmic}
\end{algorithm}


To evaluate the policy and whether it aligns with an expected value optimisation or a growth rate optimisation, we train $N$ independent agents for fixed values of the environment parameters $M$, the number of repetitions in one training episode, and $p$, the probability of the best outcome for the invested fraction in the portfolio. For a fixed set of rewards, we can predict the agent's policies resulting from either traditional expected value optimisation or the optimal policy following the Kelly criterion. These generic policies are illustrated in Figure~\ref{fig:kellygeneric}. Optimising for expected values yields an all-or-nothing strategy, as shown by the red dashed curve. Following an expected value optimisation, we invest all our wealth as soon as the probability of gaining wealth is such that the expected return exceeds 1, as indicated by the prediction $p_{E}$. The optimal policy, as derived before in Equation~\ref{eq:kelly_explicit}, is shown by the blue solid curve.

\section{Experiment results}
\subsection{Deep Q-Network and the Toy Model}
\label{sec:dqn}
\begin{figure}
  \centering
  \begin{subfigure}{0.48\textwidth}
    \includegraphics[width=\linewidth]{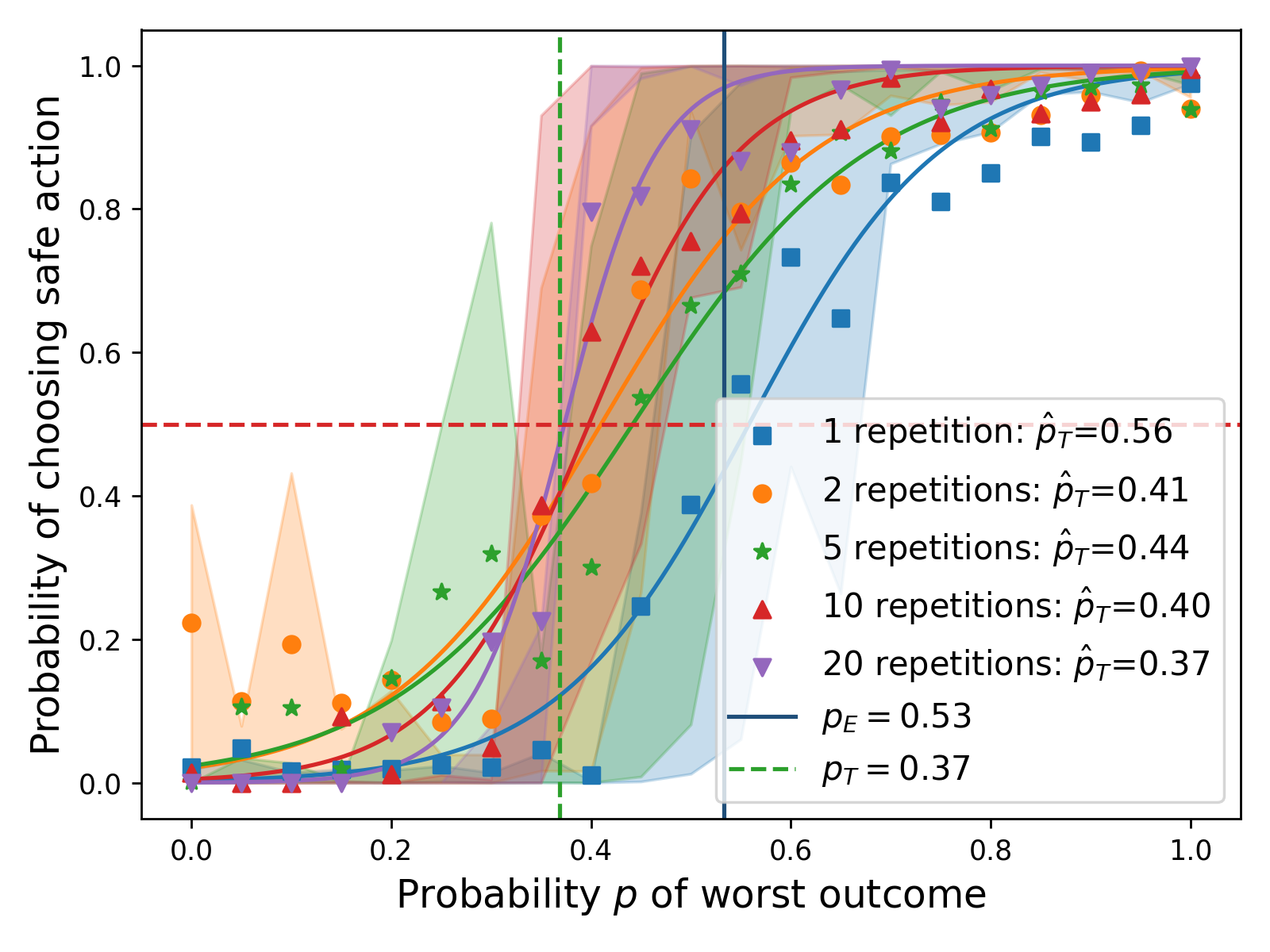}
    \caption{}
    \label{fig:DQN_toymodel}
  \end{subfigure}
  \hfill
  \begin{subfigure}{0.48\textwidth}
    \centering
    \includegraphics[width=\linewidth]{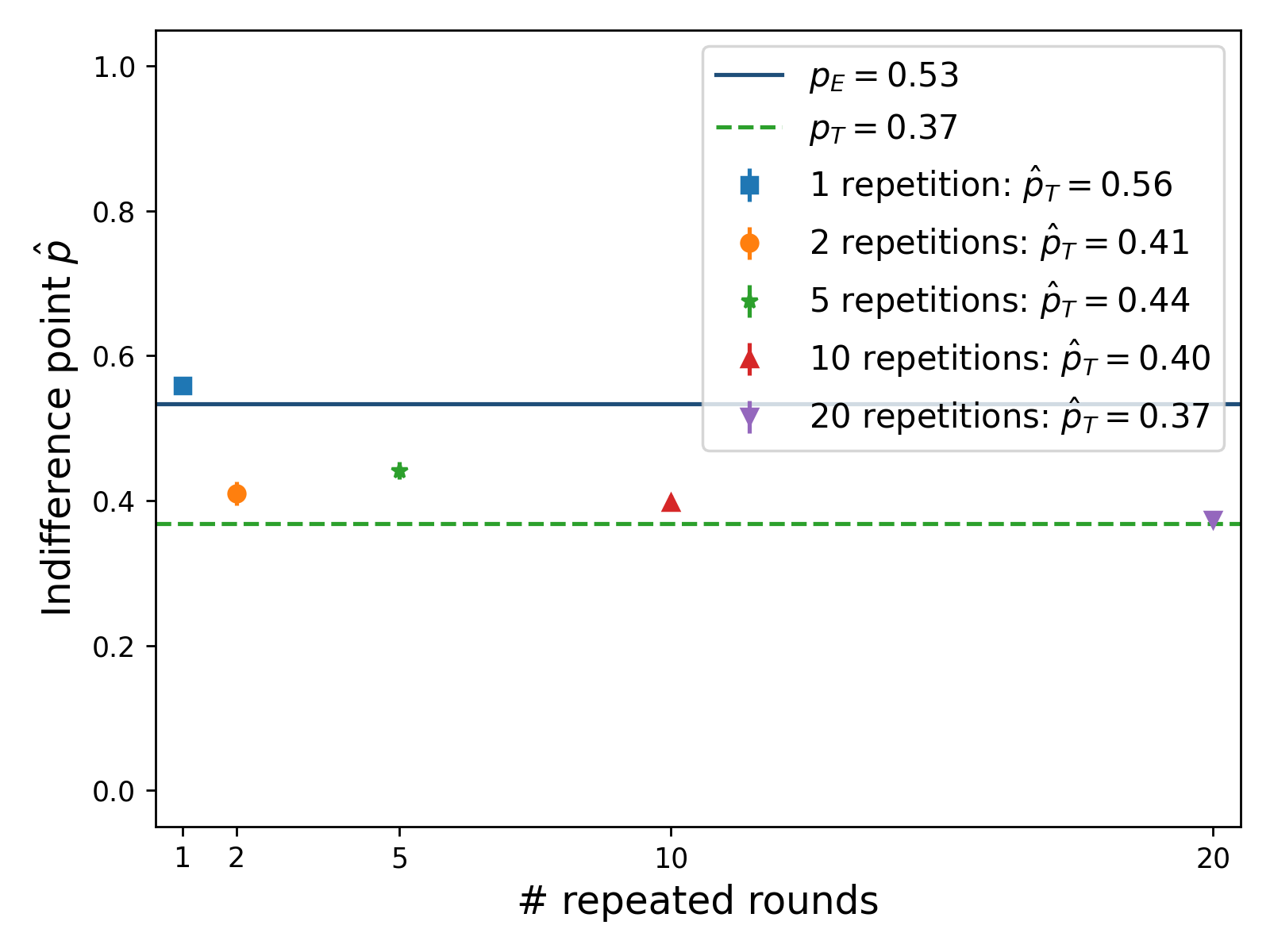}
    \caption{}
    \label{fig:DQN_policy}
  \end{subfigure}

  \caption{Learned policies under varying repetitions $M$ in the toy model. Probability of selecting the safe action as a function of $p$, shown in~\subref{fig:DQN_toymodel}, the probability of the worst risky outcome—single-step training reproduces expected-value behaviour, while repetitions expose path dependence. The indifference point $p_{0}$ shifts from the expected-value prediction $p_{E}$ (blue, solid) towards the time-growth prediction $p_{T}$ (green, dashed) as $M$ increases~\subref{fig:DQN_policy}.}
  \label{fig:drl_toymodel}
\end{figure}
To examine how reinforcement-learning agents behave in the simplified environment, as described in Section~\ref{sec:toymodel}, we train agents on the toy model problem to optimise their decisions between the safe and risky actions. Each agent is trained for a specific fixed value for the probability of receiving the worst outcome on the risky bet, $p$, and a number of repetitions of the decision in one training iteration, $M$. 

In Figure~\ref{fig:DQN_toymodel}, we show the resulting policies averaged across all agents trained with a specific set of environment variables ($p$, $M$). These results clearly illustrate the effect of the expected value at the centre of the update formalism. When agents are trained in a traditional context with a single decision per iteration, the policy resembles a perfect expected-value optimiser. Similar to traditional update rules, we can make the agent context-aware by including time as an explicit feature in the training, tracing agent trajectories across multiple repetitions of the same decision. With the inclusion of the function approximator in our DQN setup, we find that evolution from traditional policy optimisation based on expected values occurs rather quickly, even with a simple implementation using a single hidden layer. 

In Figure~\ref{fig:DQN_policy}, we show the effect on the indifference point. The location of this point indicates the underlying dynamic that the agent optimises for. The agents trained on the single decision closely resemble the estimate for a perfect expected value policy. However, as soon as we start repeating the process over the time dimension, the agents quickly learn to optimise for time growth and align with the estimate from a time-dependent dynamic. These findings highlight that increasing architectural complexity alone cannot enable reinforcement learning to make appropriate decisions in a non-ergodic context. The agent must either be informed explicitly, through transformed rewards or objective functions, or implicitly, by introducing temporal repetitions into the learning setup.

\subsection{Actor-Critic Model and Portfolio Assignment}
\label{sec:acKelly}
\begin{figure}
  \centering
  \begin{subfigure}{0.48\textwidth}
    \centering
    \includegraphics[width=\linewidth]{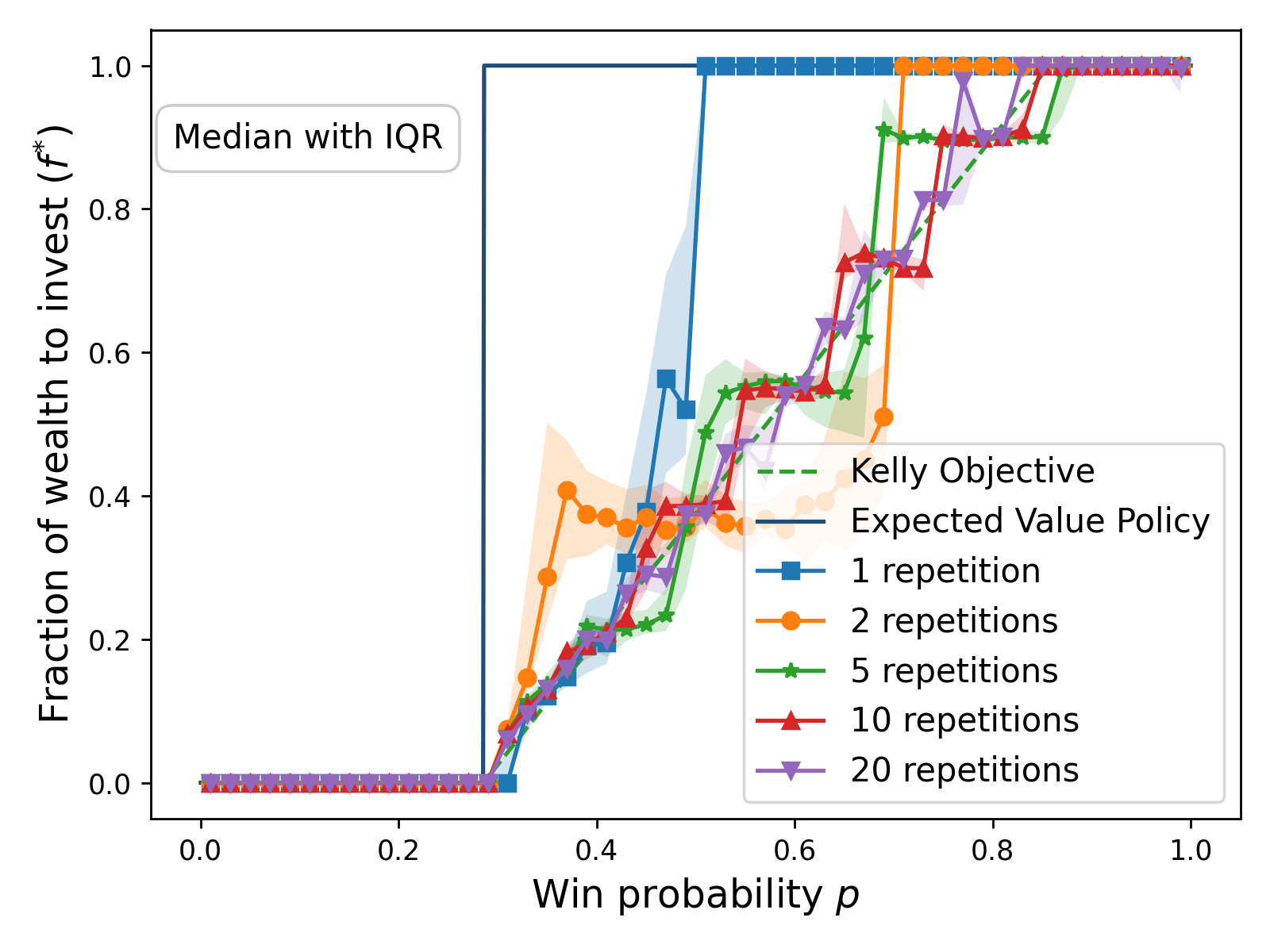}
    \caption{}
    \label{kelly_mean}
  \end{subfigure}
    \hfill
  \begin{subfigure}{0.48\textwidth}
    \centering
    \includegraphics[width=\linewidth]{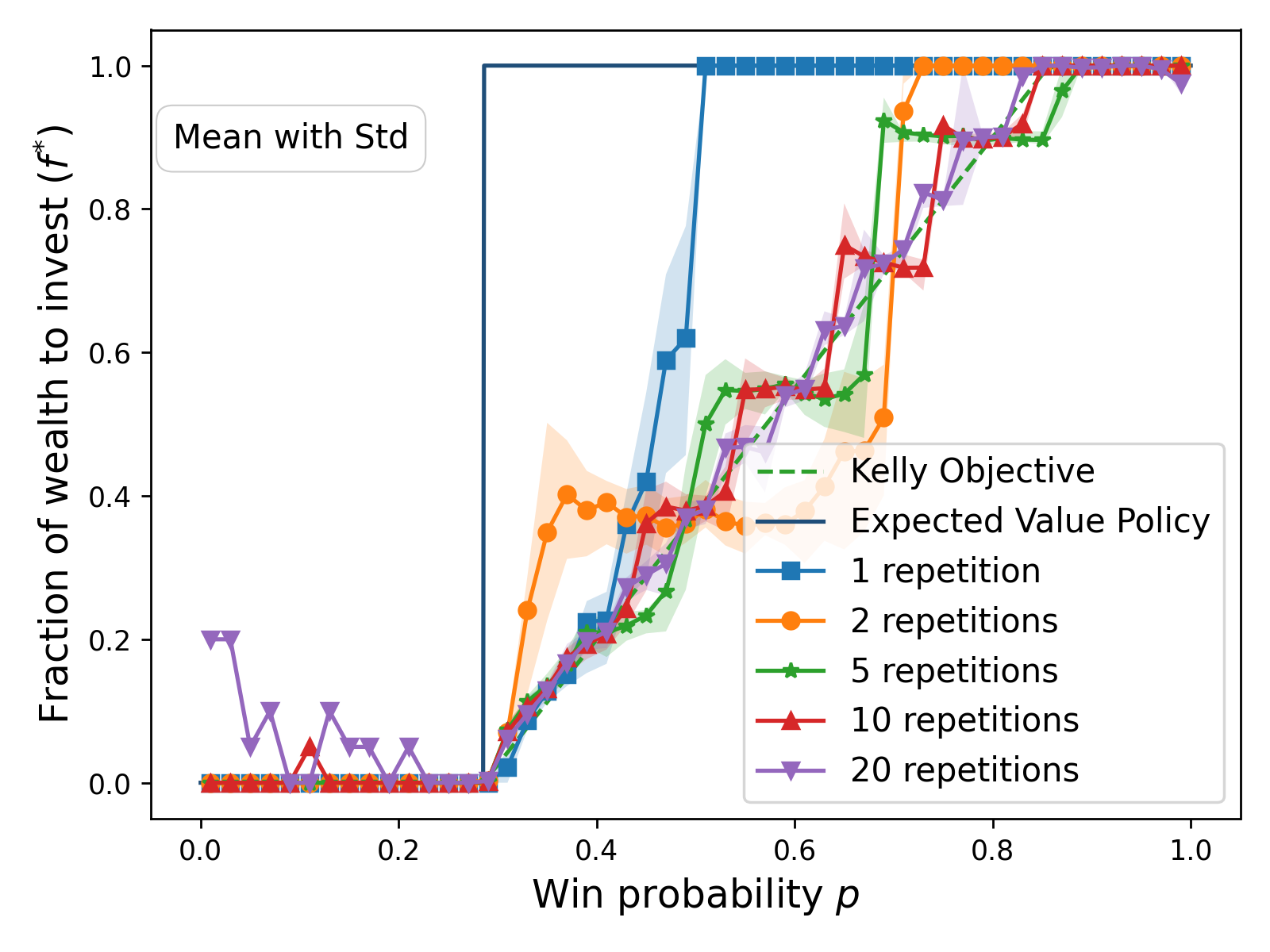}
    \caption{}
    \label{kelly_median}
  \end{subfigure}

  \caption{Resulting policy for an actor-critic DRL model trained on optimising the portfolio assignment problem. The mean policy for agents with identical training parameters is shown in~\subref{kelly_mean}, the median of these policies is shown in~\subref{kelly_median}. The agents only learn the optimal strategy prescribed by the Kelly Objective when subjected to path-dependent training, as suggested by our alternative training method.}
  \label{fig:kellypolicy}
\end{figure}

\begin{figure}
  \centering
  \begin{subfigure}{0.48\textwidth}
    \centering
    \includegraphics[width=\linewidth]{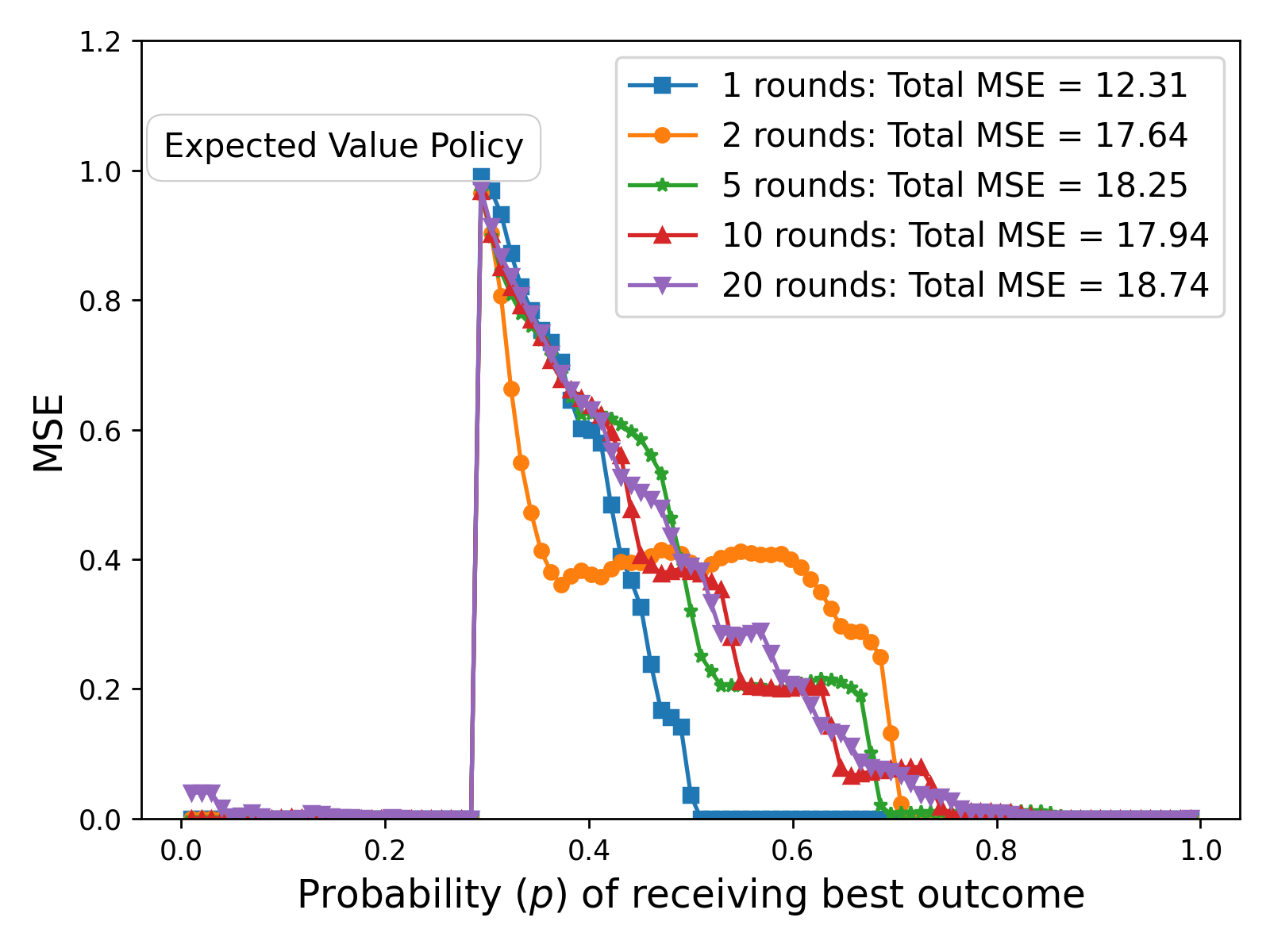}
    \caption{}
    \label{ev_MSE}
  \end{subfigure}
    \hfill
  \begin{subfigure}{0.48\textwidth}
    \centering
    \includegraphics[width=\linewidth]{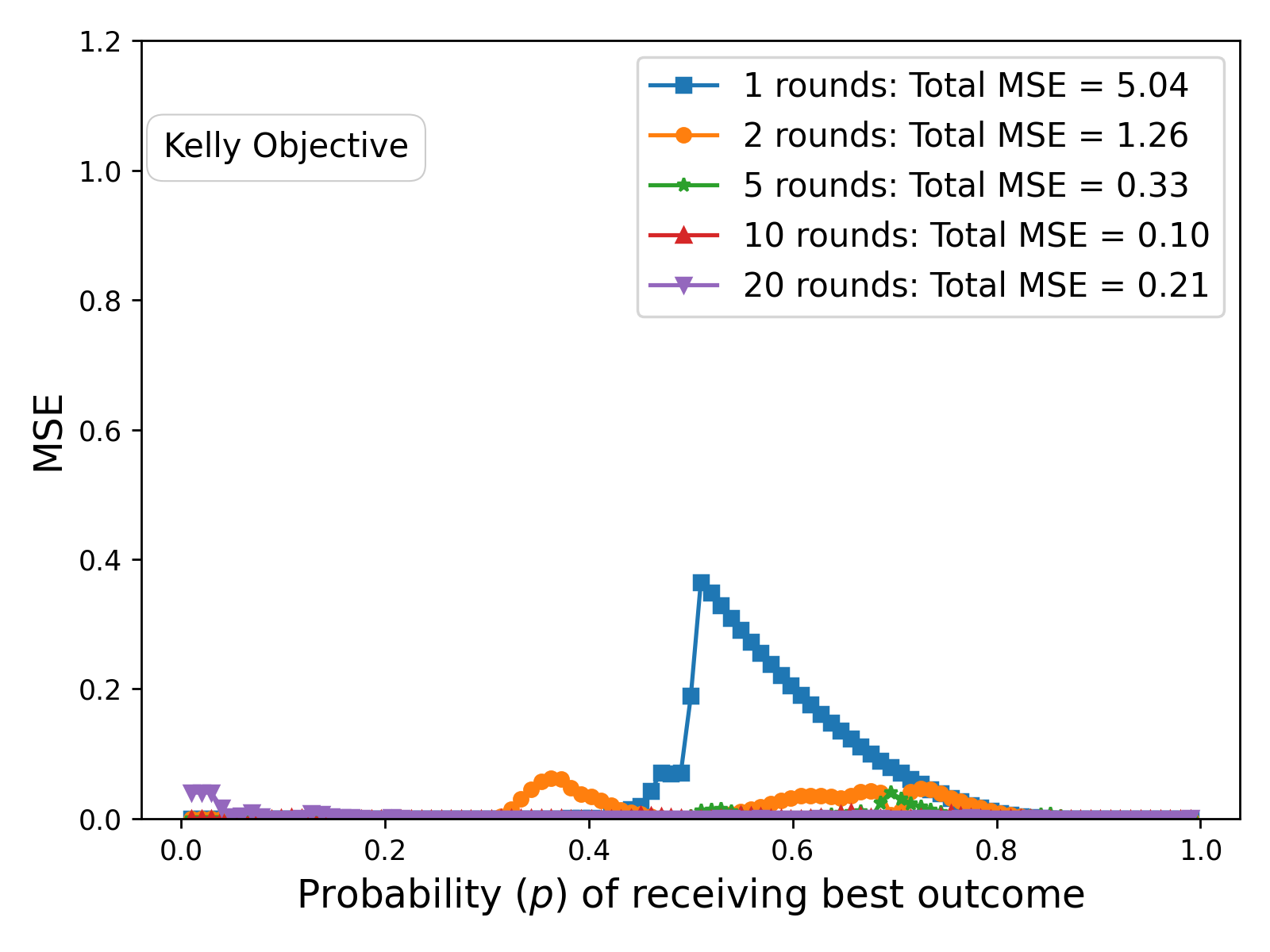}
    \caption{}
    \label{kelly_MSE}
  \end{subfigure}

  \caption{Analysis of the mean squared error (MSE) between the different policies and the theoretical policies based on the expected value optimisation~\subref{ev_MSE}, and between the policies and the optimal Kelly Objective~\subref{kelly_MSE}. The total MSE over the full policy is shown in the legend for each number of repetitions, indicating that, as the number of repetitions increases, the policy aligns more closely with the optimal policy.}  
  \label{fig:kelly_mse}
\end{figure}

The optimal policy for an agent trained on the portfolio assignment problem from Section~\ref{sec:pamodel} is described by the Kelly objective. An agent which optimises for expected values of its wealth, under an ergodic assumption, will result in a suboptimal strategy based on the prediction $p_{E}$, as derived in Equation~\ref{eq:pa_exp}. A pure policy, based on expected value or the Kelly objective, can be plotted as a function of the portfolio's reward choices. When trained with an initial wealth $W_{0} = 10$ and fixed values of the probability $p$ of gaining from the portfolio, our agents confirm once more their reliance on the expected value for optimisation. In Figure~\ref{fig:kellypolicy}, we show the resulting policy for different repetitions, and we confirm that the traditional training on this problem would result in a policy with close similarity to the expected value prediction. However, in this advanced setting, our actor-critic solution enables the agent to achieve near-perfect alignment with the optimal policy under the Kelly criterion. To illustrate alignment with the optimal policy, we can compare the mean-squared errors of each agent's predictions to the theoretical policies. These results are shown in Figure~\ref{fig:kelly_mse}. We find that our solution again enables the agent to account for the path-dependent context, and that the alignment with the optimal policy remains strong even with a relatively small number of repetitions.

An alternative approach to training agents for the portfolio assignment problem involves full policy learning. In previous results, agents are trained in a context with a fixed probability of gaining from the portfolio. In contrast, to learn the full policy for an individual agent, we randomly sample values of p for each new episode.
We note that full policy learning requires an agent to train over longer episodes to learn the full range of the probability distribution and its related optimal fraction. We present the results of an initial training setup for policy learning on the portfolio assignment problem in Appendix~\ref{app:kelly_policy}.

\section{Conclusion}
Optimising expected value is deeply embedded in reinforcement learning. We have shown that using a more advanced technique, which implements a function approximation of the value function, does not resolve the problematic reliance of an expected return in a deep RL formulation of a multiplicative dynamic. Our results show that, in a traditional implementation of a Deep Q-network for a simple toy model of a multiplicative reward setting, an agent retains its focus on expected value rather than learning an optimal policy based on growth rates. We further demonstrate that the proposed solution, which explicitly embeds the temporal dynamics via repeated action selection, enables the agent to be informed of the path-dependent nature of the underlying dynamics. These results generalise the solution proposed in~\cite{verbruggen2025reinforcement}, which challenges the essential role of expected-value optimisation. Using a DQN, we find that evolution to the optimal policy can occur with fewer repetitions in the toy model. 

In an additional experiment, we extend the context to a continuous action space, allowing the agent to select the optimal fraction of initial wealth to invest in a portfolio assignment problem. Our results confirm that extending this problem to an actor-critic setting suffers from the same issue when implemented using traditional methods. Our proposed solution confirms that, by explicitly enforcing path dependence, an agent can transition to the optimal policy and align with the Kelly criterion. 

Although the studies reported in this work focus on two reference problems in a simplified context, the results confirm the hypothesis. A traditional implementation of RL agents, whether using discrete update rules or Deep RL methods with value function approximation and continuous action spaces, suffers from a lack of information about temporal dynamics. The proposed solution, which uses repeated exposure of the agent to the effect of its decisions over extended time horizons, remains effective in more advanced contexts as well. This work can inspire solutions for effective policy learning in contexts where the ergodic assumption breaks down and provides an elegant conceptual framework to encompass essential considerations when modelling agents in an unknown environment. The essential difference from solutions proposed in the literature for similar problems relies on the understanding of the dynamics as a process of temporal dynamics, which can be introduced during training, without an explicit change to the value function or reward signal. A future consideration of our methodology can extend to the field of risk-sensitive RL as well. 

\bibliographystyle{RS}

\begin{thebibliography}{99}

\bibitem{sutton2018reinforcement}
Sutton RS, Barto AG. 2018 {\em Reinforcement learning: An introduction}.
MIT press.

\bibitem{rummery1994line}
Rummery GA, Niranjan M. 1994 {\em On-line Q-learning using connectionist systems} vol.~37.
University of Cambridge, Department of Engineering Cambridge, UK.

\bibitem{watkins1992q}
Watkins CJ, Dayan P. 1992  Q-learning. {\em Machine learning} \textbf{8}, 279--292.

\bibitem{verbruggen2025reinforcement}
Verbruggen B, Vanhoyweghen A, Ginis V. 2025  Reinforcement learning in path-dependent, non-ergodic contexts. In {\em Proceedings A} vol. 480-2314 p. 20240415. The Royal Society.

\bibitem{peters2019ergodicity}
Peters O. 2019  The ergodicity problem in economics. {\em Nature Physics} \textbf{15}, 1216--1221.

\bibitem{peters2016evaluating}
Peters O, Gell-Mann M. 2016  Evaluating gambles using dynamics. {\em Chaos: An Interdisciplinary Journal of Nonlinear Science} \textbf{26}.

\bibitem{hulme2023reply}
Hulme O, Vanhoyweghen A, Connaughton C, Peters O, Steinkamp S, Adamou A, Baumann D, Ginis V, Verbruggen B, Price J et~al.. 2023  Reply to" The Limitations of Growth-Optimal Approaches to Decision Making Under Uncertainty". {\em Econ Journal Watch} \textbf{20}, 335--348.

\bibitem{meder2021ergodicity}
Meder D, Rabe F, Morville T, Madsen KH, Koudahl MT, Dolan RJ, Siebner HR, Hulme OJ. 2021  Ergodicity-breaking reveals time optimal decision making in humans. {\em PLOS Computational Biology} \textbf{17}, e1009217.

\bibitem{vanhoyweghen2022influence}
Vanhoyweghen A, Verbeken B, Macharis C, Ginis V. 2022  The influence of ergodicity on risk affinity of timed and non-timed respondents. {\em Scientific Reports} \textbf{12}, 3744.

\bibitem{vanhoyweghen2023human}
Vanhoyweghen A, Ginis V. 2023  Human decision-making in a non-ergodic additive environment. {\em Proceedings of the Royal Society A} \textbf{479}, 20230544.

\bibitem{VANHOYWEGHEN2025101663}
Vanhoyweghen A, Ginis V, Macharis C. 2025  Beyond expected travel time: Unveiling the role of ergodicity breaking in mobility decisions and effective transportation policy. {\em Transportation Research Interdisciplinary Perspectives} \textbf{34}, 101663.
(\href{http://dx.doi.org/https://doi.org/10.1016/j.trip.2025.101663}{https://doi.org/10.1016/j.trip.2025.101663})

\bibitem{peters2023insurance}
Peters O. 2023  Insurance as an ergodicity problem. {\em Annals of Actuarial Science} \textbf{17}, 215--218.

\bibitem{peters2022ergodicity}
Peters O, Adamou A. 2022  The ergodicity solution of the cooperation puzzle. {\em Philosophical Transactions of the Royal Society A} \textbf{380}, 20200425.

\bibitem{vanhoyweghen2025redistributive}
Vanhoyweghen A, Macharis C, Vincent G. 2025  Investigating the origins of redistributive behaviour in non-ergodic contexts. {\em Proceedings of the Royal Society A: Mathematical, Physical and Engineering Sciences}.
In press (\href{http://dx.doi.org/10.1098/rspa.2025.0598}{10.1098/rspa.2025.0598})

\bibitem{fant2023stable}
Fant L, Mazzarisi O, Panizon E, Grilli J. 2023  Stable cooperation emerges in stochastic multiplicative growth. {\em Physical Review E} \textbf{108}, L012401.

\bibitem{baumann2023reinforcement}
Baumann D, Noorani E, Price J, Peters O, Connaughton C, Sch{\"o}n TB. 2023  Reinforcement learning with non-ergodic reward increments: robustness via ergodicity transformations. {\em arXiv preprint arXiv:2310.11335}.

\bibitem{sheng2025beyond}
Sheng X, Baumann D. 2025  Beyond expected value: geometric mean optimization for long-term policy performance in reinforcement learning. {\em arXiv preprint arXiv:2508.21443}.

\bibitem{howard1972risk}
Howard RA, Matheson JE. 1972  Risk-sensitive Markov decision processes. {\em Management science} \textbf{18}, 356--369.

\bibitem{heger1994consideration}
Heger M. 1994  Consideration of risk in reinforcement learning. In {\em Machine Learning Proceedings 1994} , . Elsevier.

\bibitem{mihatsch2002risk}
Mihatsch O, Neuneier R. 2002  Risk-sensitive reinforcement learning. {\em Machine learning} \textbf{49}, 267--290.

\bibitem{di2012policy}
Di~Castro D, Tamar A, Mannor S. 2012  Policy gradients with variance related risk criteria. {\em arXiv preprint arXiv:1206.6404}.

\bibitem{haarnoja2018soft}
Haarnoja T, Zhou A, Hartikainen K, Tucker G, Ha S, Tan J, Kumar V, Zhu H, Gupta A, Abbeel P et~al.. 2018  Soft actor-critic algorithms and applications. {\em arXiv preprint arXiv:1812.05905}.

\bibitem{prashanth2016variance}
Prashanth L, Ghavamzadeh M. 2016  Variance-constrained actor-critic algorithms for discounted and average reward MDPs. {\em Machine Learning} \textbf{105}, 367--417.

\bibitem{bielecki1999risk}
Bielecki TR, Pliska SR. 1999  Risk-sensitive dynamic asset management. {\em Applied mathematics and optimization} \textbf{39}, 337--360.

\bibitem{bellemare2017distributional}
Bellemare MG, Dabney W, Munos R. 2017  A distributional perspective on reinforcement learning. In {\em International conference on machine learning} pp. 449--458. PMLR.

\bibitem{biswas2023ergodic}
Biswas A, Borkar VS. 2023  Ergodic risk-sensitive control—a survey. {\em Annual Reviews in Control} \textbf{55}, 118--141.

\bibitem{redner1990random}
Redner S. 1990  Random multiplicative processes: An elementary tutorial. {\em Am. J. Phys} \textbf{58}, 267--273.

\bibitem{mnih2015human}
Mnih V, Kavukcuoglu K, Silver D, Rusu AA, Veness J, Bellemare MG, Graves A, Riedmiller M, Fidjeland AK, Ostrovski G et~al.. 2015  Human-level control through deep reinforcement learning. {\em nature} \textbf{518}, 529--533.

\bibitem{kelly1956new}
Kelly JL. 1956  A new interpretation of information rate. {\em The Bell System Technical Journal} \textbf{35}, 917--926.

\bibitem{lillicrap2015continuous}
Lillicrap TP, Hunt JJ, Pritzel A, Heess N, Erez T, Tassa Y, Silver D, Wierstra D. 2015  Continuous control with deep reinforcement learning. {\em arXiv preprint arXiv:1509.02971}.

\end{thebibliography}

\appendix
\section{Appendix}
\subsection{DQN model parameters}
\label{appendix_dqn}
The Deep Q-network used in the experiment is configured as a standard, shallow, fully connected neural network. Implementing this model in the RL training routine can be done in a traditional integration. The neural network receives one input value representing the wealth at the current timestep and generates an output on two neurons. The network's prediction provides an estimate of the agent's preference for each of the actions it can take, either safe or risky. In Section~\ref{sec:dqn} we introduced our experiment using a standard DQN implementation on the toy problem. In Table~\ref{table_dqnparameters}, we present the neural network model parameters and the training parameters used in the final experiments.

\begin{table}
    \caption{Parameter values for the training of the DQN model on the toy problem}
    \centering
    \begin{tabular}{|l|c|l|c|}
        \hline
        \textbf{Parameter} & \textbf{Value} & \textbf{Parameter} & \textbf{Value} \\
        \hline
        $r_{1}$ & $0.5$ & $\#$Episodes & $1 \times 10^{4}$ \\
        \hline
        $r_{2}$ & $2.0$ & $\#$Agents & $40$ \\
        \hline
        $r_{3}$ (safe multiplier) & $1.2$ & Discount $\gamma$ & $0.9$ \\
        \hline
        $\epsilon$-greedy strategy & Decaying & $\epsilon$ initial & $1.0$ \\
        \hline
        $\epsilon$ threshold & $0.05$ & $\epsilon$ decay rate & $0.995$ \\
        \hline
        Optimizer & Adam & $\alpha$ (lr) & $0.8$ \\
        \hline
        Loss function & Smooth L1 & Batch size & $2$ \\
        \hline
        Hidden layers & 1 & Hidden width $H$ & $16$ \\
        \hline
        Activation & ReLU & Output dimension & $2$ \\
        \hline
    \end{tabular}
    \label{table_dqnparameters}
\end{table}

\subsection{Actor-Critic Model and Training Parameters}
\label{appendix_drlkelly}
The portfolio assignment problem involves an agent selecting the proper fraction of its wealth ($f$) to allocate to a stochastic bet. In our deep RL implementation, we use an actor-critic model to enable the actor to predict the optimal fraction ($f^{*}$) from a continuous action space and to use the environment to calculate the new wealth ($W_{t+1}$) based on its prediction. The critic is trained on the agent's past trajectories, gathering states, actions, and rewards to learn a value function estimate. The parameters for the actor-critic model and the agents' training are shown in Table~\ref{table_ac_kelly_cont}. We also present a more detailed discussion of the model design and implementation for agent training, following the condensed pipeline from Section~\ref{sec:acKelly}. 
\begin{table}
    \caption{Parameter values for the Actor--Critic portfolio model (Kelly objective, continuous action)}
    \centering
    \begin{tabular}{|l|c|l|c|}
        \hline
        \textbf{Parameter} & \textbf{Value} & \textbf{Parameter} & \textbf{Value} \\
        \hline
        $r_{1}$ & $3.0$ & \#Episodes & $1 \times 10^5$ \\
        \hline
        $r_{2}$ & $0.2$ & \#Agents & $20$ \\
        \hline
        $r_{\text{safe}}$ & $1.0$ & Rounds/episode $M$ & $(1,2,5,10,20)$ \\
        \hline
        Optimizer & Adam & Action space & $f\in[0,1]$ \\
        \hline
        Policy LR $\eta_{\pi}$ & $1\times10^{-3}$ & Value LR $\eta_{V}$ & $1\times10^{-3}$ \\
        \hline
        Hidden layers actor & 1  & Hidden width actor & 32  \\
        \hline
        Hidden layers critic & 1  & Hidden width critic & 32  \\
        \hline
        Activation & ReLU & Critic loss & Smooth L1 \\
        \hline
    \end{tabular}
    \label{table_ac_kelly_cont}
\end{table}

\subsection{Policy training result for AC model}
\label{app:kelly_policy}
\begin{figure}
    \centering
    \includegraphics[width=\linewidth]{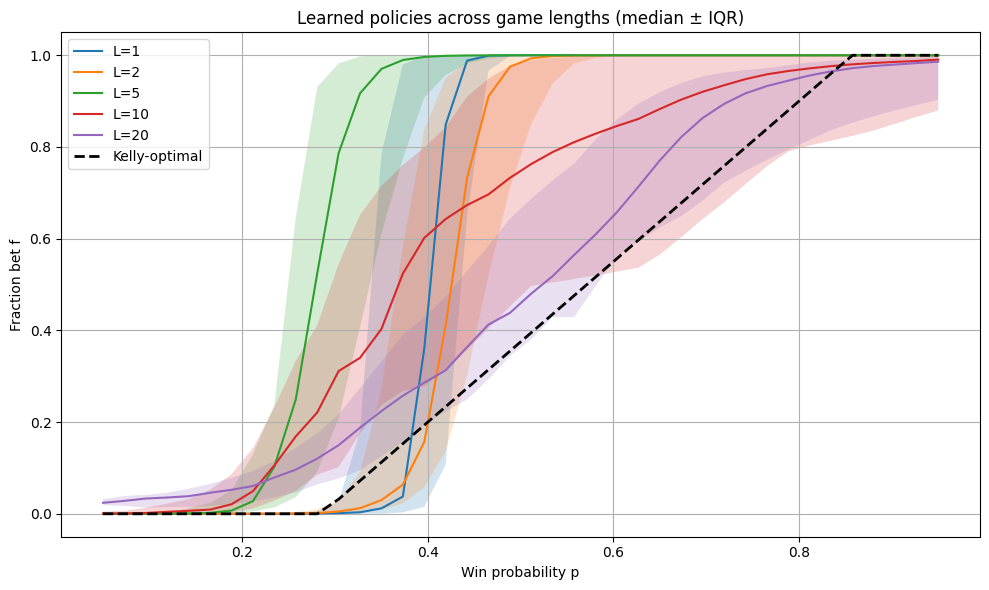}
    \caption{Policies learned by agents trained using an Actor-Critic model with randomly selected values for parameter $p$, the probability to gain from the investment in the portfolio. This result illustrates that agents can even learn full policies, and within reasonable precision for the portfolio assignment problem.}
    \label{fig:policy_AC}
\end{figure}

In addition to training individual agents for fixed values of p, we can also train a single actor–critic model that develops a complete policy across the entire range of win probabilities. Figure~\ref{fig:policy_AC} shows the resulting learned policies for different repetitions $M$. Each curve represents the median betting fraction f across agents, with shaded regions indicating the interquartile range (IQR). As the number of repeated rounds increases, the learned policy gradually transitions from an expected-value strategy, underestimating the optimal bet, to the Kelly-optimal solution. This confirms that the actor–critic architecture can generalise over all probabilities $p$ to gain from the portfolio and approximate the full functional form of the optimal policy when trained in a path-dependent setting.

However, this global policy-learning task is considerably more difficult to stabilise. Unlike the fixed-$p$ training runs, where convergence occurs relatively quickly, the joint optimisation across all probabilities requires both a larger number of training episodes and longer game lengths to converge.

\end{document}